# CT-IDP: Segmentation-Derived Quantitative Phenotypes for Interpretable Abdominal CT Disease Classification

Lavsen Dahal[1,3], Joseph Lo[1,2,3,4]

1. Center for Virtual Imaging Trials, Carl E. Ravin Advanced Imaging Laboratories,

2. Department of Radiology, Duke University School of Medicine, Durham, NC, USA

3. Electrical and Computer Engineering, Pratt School of Engineering, Duke University, Durham, NC, USA

4. Medical Physics Graduate Program, Duke University, Durham, NC, USA

**Abstract**

In this retrospective multi-institutional study, a quantitative phenotyping framework, CT-IDP (CT Image-Derived Phenotypes) was developed on the MERLIN abdominal CT benchmark (training, validation, and test sets: 15,175, 5,018, and 5,082 studies, respectively) and externally evaluated on two independent cohorts: a private institutional dataset (n = 2,000) and AMOS (n = 1,107). Multi-organ segmentations were generated with TotalSegmentator and used to derive over 900 organ and compartment-level descriptors spanning morphometry, attenuation, and contextual/burden findings. Sparse disease-specific logistic regression with elastic-net regularization was trained on MERLIN and externally validated under a frozen specification. Performance was compared against a DINOv3-based vision-transformer baseline using AUC and average precision (AP), supported by phenotype-stratified audits and coefficient-level inspection. Macro-AUC for CT-IDP versus the baseline was 0.897 (95% CI, 0.889–0.904) versus 0.880 (0.871–0.889) on MERLIN, 0.877 (0.864–0.889) versus 0.857 (0.845–0.869) on the private institutional cohort, and 0.780 (0.763–0.796) versus 0.756 (0.739–0.772) on AMOS.

## Introduction

Abdominal CT volumes continue to rise, intensifying the need for automated computer-aided diagnosis methods that perform reliably across scanners, protocols, and institutions [1-3]. Existing abdominal CT disease classifiers, spanning Convolutional Neural Networks (CNNs), vision transformers, and vision-language models, primarily derive representations from image appearance [4-7]. Although these models can achieve strong benchmark performance, they may remain vulnerable to shortcut learning, relying on texture-dominant rather than stable quantitative manifestations of disease [8, 9].

This limitation is most consequential for findings whose radiologic definitions are themselves measurements: organ size, vessel caliber, tissue attenuation, cyst burden, or calcific burden, where disease is expressed through measurable anatomy rather than local texture [10-12]. We hypothesized that representations built directly from such measurements would be both more interpretable and more robust under dataset shift than appearance-based embeddings.

Quantitative imaging is not a new idea. Traditional radiomics has typically been disease or lesion-centric: a region of interest is segmented, and a large, often texture-dominant feature set is computed within it, which can be sensitive to reconstruction kernel, noise, and scanner variation [13-17]. Recent advances in multi-organ CT segmentation [18-21] enable a different formulation, in which a single whole-body segmentation supports a shared descriptor library that is computed once per case and applied across multiple disease classifiers. The library combines Image Biomarker Standardisation Initiative (IBSI) aligned organ-level summaries with cross-organ and compartment-level composites.

In this work, we present **CT-IDP (CT image-derived phenotypes)**, a phenotyping framework that operationalizes this anatomy-centric formulation for interpretable multi-label abdominal CT classification. Rather than replacing learned image representations, CT-IDP provides a measurement-grounded alternative: sparse disease-specific models trained on standardized organ-level descriptors yield explicit, auditable prediction rules in which every output is traceable to a concrete anatomical measurement. This supports mechanistic inspection: external performance differences can be traced to specific coefficients and the dataset-level statistics that drive them and characterizes an operating regime in which measurement-grounded models suffice and one in which appearance-based representations remain necessary.

We evaluated CT-IDP on the public MERLIN benchmark [22] for development and on two independent external datasets - the public AMOS dataset [23] and a private institutional dataset for frozen external testing. To contextualize the method against learned image representations, we compared it with a DINOv3-based vision-transformer baseline [24] trained on the same development data.

# Materials and Methods

## Study Design and Datasets

This retrospective, multi-institutional study evaluated CT-IDP, a segmentation-derived quantitative phenotyping framework for multi-label abdominal CT classification (Table 1). The study was approved by the local Institutional Review Board with waiver of informed consent.

**Development Dataset.** The public MERLIN dataset was used for model development and internal testing under its official training, validation, and test splits. Of MERLIN's 30 study-level findings, 22 were retained for evaluation, the subset for which anatomically indexed quantitative phenotypes were predefined.

**External Test Datasets.** Two independent datasets were used for frozen external evaluation under the locked CT-IDP specification: an institutional abdominal CT dataset retrospectively sampled from archival examinations with associated radiology reports (21 evaluable labels), and the public AMOS dataset (8 evaluable labels). Because evaluable label sets differed across datasets, macro-averaged performance is reported within each dataset's own label set; an 8-label subset shared by all three datasets was used for direct cross-dataset comparison.

## Reference Labels

MERLIN labels were used as published. For the private institutional dataset, study-level labels were derived from radiology reports by two independent LLM-assisted NLP pipelines based on MedGemma [25] and Qwen[26] using the RATE framework released with Pillar-0 [7]: each pipeline extracted findings from the report and mapped them to the study label space. Concordant positive or negative assignments were accepted as the final label; disagreements or absent report evidence were marked uncertain and excluded from the affected per-finding metric. For AMOS, evaluation was restricted to the 8 labels that could be unambiguously mapped to the study task definition and supported by the phenotype library.

## CT-IDP framework

Multi-organ segmentations were generated with TotalSegmentator and mapped to the native CT image grid before descriptor extraction. Descriptors were computed in native image space using voxel-spacing metadata so that organ geometry, attenuation, and burden estimates were expressed in physical units. The resulting library comprised more than 900 organ and compartment-level descriptors spanning three families: morphometry, attenuation, and contextual/burden as summarized in Table 2. Full descriptor definitions and computation code are publicly documented at https://phenoct.github.io/.

## Sparse disease-specific modeling

For each target finding, a disease-specific classifier was trained on the descriptor library. Missing values, primarily from structures outside the scan field of view, were imputed using MERLIN-dataset summary statistics; descriptors were z-score standardized using MERLIN parameters; the same imputation and scaling rules were applied unchanged at external evaluation. Highly correlated descriptors were removed by pairwise filtering ($|r| > 0.95$).

Features were selected by stratified 5-fold cross-validation under elastic-net logistic regression, sweeping regularization strength ($C \in \{0.001, 0.01, 0.1, 1.0\}$) and the L1/L2 mixing ratio (l1_ratio ∈ {0.1, 0.5, 0.9}) and optimizing average precision; features retained in at least 60% of folds were kept. The sweep was used to characterize sensitivity of selection to regularization rather than to tune per-disease hyperparameters, and the locked classifier was fit at a fixed central operating point (C = 0.1, l1_ratio = 0.5) to avoid overfitting on the small positive counts of several findings. Models were trained on the selected feature set with balanced class weighting [27]. Classifiers were trained independently per finding; after development, the full specification - retained descriptors, imputation rules, scaling parameters, and coefficients was locked and applied identically to both external datasets. All models were implemented in scikit-learn v1.5.1 (Python 3.12).

## Vision-Transformer Baseline

As a deep learning benchmark, we used a DINOv3-based vision transformer initialized from self-supervised pretrained weights [24]. CT volumes were intensity-clipped to [−1000, 1000] HU, resampled to 1.5 × 1.5 × 3.0 mm, and center-cropped or zero-padded to 224 × 224 × 160. Each axial position was encoded using a tri-slice scheme - central slice plus immediate superior and inferior slices providing limited through-plane context while retaining compatibility with 2D pretrained weights. Slice-level embeddings were aggregated to study-level predictions via mean pooling over the volume before the multi-label classification head. This baseline was intended as a standardized learned-image reference model trained under the same development setting, rather than an exhaustive search over deep architecture, resolutions, or aggregation mechanisms.

The model was trained on the MERLIN development split with physiologically constrained augmentation (horizontal flipping disabled to preserve organ laterality). Missing study-level labels were initially excluded from the loss and later incorporated with reduced weight via a dynamic uncertainty curriculum. Optimization used grouped learning rates (backbone vs heads) with cosine decay over 20 epochs. Model selection used MERLIN validation macro-AUC; the locked model was then evaluated on the MERLIN internal test set and both external datasets under the same frozen-evaluation protocol as CT-IDP.

## Evaluation and Statistical Analysis

Performance was assessed by AUC and average precision (AP), with 95% confidence intervals from 2,000 nonparametric bootstrap replicates. Macro-metrics were unweighted means across each dataset's evaluable label set. Per-finding ΔAUC and ΔAP between CT-IDP and the ViT baseline were estimated by paired bootstrap. Phenotype-stratified analyses were performed for representative findings (gallstones, hepatic steatosis, AAA, renal cyst) to characterize performance as a function of phenotype magnitude. For findings with notable external gaps, retained coefficients were inspected and cross-referenced against training-label co-occurrence to surface candidate failure mechanisms.

# Results

## Overall benchmark performance across internal and external datasets

CT-IDP outperformed the DINOv3 baseline on macro-AUC and macro-AP across all three datasets under the locked frozen-evaluation protocol (Table 3). Because evaluable label sets differed by dataset, we also computed performance on the 8-label subset shared by all three: macro-AUC was 0.884 vs 0.865 (DINOv3) on MERLIN, 0.864 vs 0.841 on the institutional dataset, and 0.780 vs 0.756 on AMOS, confirming that the dataset-level advantage was not an artifact of label-set composition.

A consistent regime pattern emerged on the harmonized 8-finding set (Table 4): CT-IDP gains were largest and most stable on findings whose radiologic definition is itself a measurement: gallstones, hepatic steatosis, atherosclerosis, and splenomegaly while ascites and bowel obstruction, which depend on compartmental fluid distribution and bowel configuration, consistently favored DINOv3. Renal cysts and hydronephrosis showed mixed cross-dataset behavior driven by lesion observability, examined in the phenotype-stratified analysis below.

## Internal label-level performance on MERLIN

Across the 22 MERLIN findings (Figure 2A), ΔAUC favored CT-IDP for nearly all morphometry and attenuation findings and favored ViT for most contextual/burden findings. The largest CT-IDP gains were on hepatic steatosis (+0.156), abdominal aortic aneurysm (+0.141), gallstones (+0.111), and pancreatic atrophy (+0.099); the largest losses on free air (−0.102), hydronephrosis (−0.085), and bowel obstruction (−0.080). Family-level mean ΔAUC was +0.050 for morphometry, +0.054 for attenuation, and −0.034 for contextual/burden, with non-overlapping bootstrap 95% CIs separating contextual/burden from the other two (Figure 2B). Ceteris paribus model-response curves for representative retained features were monotonic and aligned with physiologic expectation predicted probability rose with aortic diameter, liver-to-body ratio, and gallbladder maximum HU, and fell with the liver–spleen attenuation difference (Figure 2C) confirming that the classifier's decision rule reflects the same anatomical thresholds clinicians use to characterize these findings.

## Generalization under frozen external validation

The internal regime pattern was preserved externally (Figure 3). On the institutional dataset (Figure 3A, 21 findings), CT-IDP led most strongly on gallstones (ΔAUC +0.255), prostatomegaly (+0.149), and pancreatic atrophy (+0.121), and trailed most on free air (−0.126), bowel obstruction (−0.119), and ascites (−0.070). On AMOS (Figure 3B, 8 findings), gallstones (+0.178) and hepatic steatosis (+0.094) led, while renal cyst (−0.088) and ascites (−0.053) favored ViT. Aggregating by descriptor family (Figure 3C) reproduced the internal MERLIN pattern: morphometry and attenuation favored CT-IDP on both external datasets; contextual/burden favored ViT.

Dataset-level macro-AUC degraded modestly from MERLIN to the institutional dataset (0.897 → 0.877) and more substantially to AMOS (0.897 → 0.780), but the family-level ordering was preserved on both external datasets. The larger AMOS degradation across all three descriptor families is consistent with greater protocol heterogeneity particularly contrast phase between AMOS and MERLIN's predominantly portal venous phase distribution. Findings whose discriminative signal depends on parenchymal HU, such as renal cyst (which relies on parenchyma-to-cyst contrast), were disproportionately affected on AMOS.

One within-family exception is informative. On the institutional dataset, hepatomegaly was the single morphometry finding favoring ViT (ΔAUC −0.063). Coefficient inspection of the locked classifier showed that the third-largest weight was carried by liver–spleen ΔHU, a descriptor whose primary clinical signal is hepatic steatosis rather than liver enlargement. We hypothesize that the external gap reflects multiple contributing factors: NLP-driven training-label structure that preferentially captured hepatomegaly within reports co-mentioning steatosis (producing classifier reliance on a steatosis-specific feature), institutional-dataset case mix containing hepatic mass lesions whose volume contributions are inconsistently labeled as hepatomegaly. Quantitative attribution among these contributions is left to dedicated follow-up analysis with controlled measurement and label provenance.

## Phenotype-stratified audit of quantitative model performance

To characterize each representation's operating regime, we stratified four representative findings by the measurement that defines the disease subgroup (Figure 4). Because each stratifying variable is itself a CT-IDP descriptor, this analysis is best read as an observability audit: it characterizes the regime in which each representation can or cannot encode the abnormality.

A consistent pattern emerged across findings. CT-IDP achieved its highest discrimination where disease was clearly expressed in segmented anatomy — calcified gallstones (gallbladder max HU > 200; AUC 0.97 MERLIN, 0.86 institutional), diffuse low-attenuation steatosis (ΔHU < −10), enlarged aortic caliber, and large renal cysts (≥5 cc). Discrimination dropped sharply in subgroups where the abnormality fell below the descriptor space: non-calcified gallstones (AUC 0.55, 0.50), focal or borderline steatosis with normal global ΔHU (small subgroups, n = 6 and 11), and sub-threshold cysts (<2 cc). DINOv3 retained moderate residual discrimination in several of these regimes (non-calcified gallstones AUC 0.65, 0.62; sub-threshold cysts higher than CT-IDP), consistent with appearance cues that anatomy-level descriptors cannot encode. AAA was the cleanest CT-IDP win: aortic-diameter descriptors held discrimination against both non-ectatic and ectatic comparators (AUC 0.92, 0.95), whereas DINOv3 dropped in the ectatic comparator subgroup.

Renal cyst on AMOS was the most striking observability case (Figure 3D): both models performed poorly when all report-positive cases were retained, with AUC rising monotonically as sub-threshold cysts were excluded — from 0.56 across all positives to 0.94 for cysts ≥ 5 cc. Of 398 report-positive cysts, 248 (62%) had measured volume below 2 cc and 95 (24%) below 0.5 cc, indicating that the bottleneck is lesion observability rather than representation choice: sub-threshold cysts fall near the segmentation detection limit, and both descriptor- and image-based predictions degrade in this regime.

## Qualitative audit of representative success and failure cases

Representative true-positive and error cases (Figure 5) corresponded to the same observability limits identified in the stratified analysis: non-calcified gallstones, and renal lesions reported as "too small to characterize." One error case warrants separate comment. For hepatic steatosis, a false-negative example carried a positive report-derived label, but radiologist review suggested focal rather than diffuse steatosis with global liver attenuation in the normal range. CT-IDP's prediction reflected the global descriptor faithfully but conflicted with the binary study-level label. Such cases illustrate a broader limitation of report-derived labels for radiologically heterogeneous diseases: a single positive label can correspond to disease patterns of very different extent and observability, affecting both descriptor and image-based evaluation.

# Discussion

CT-IDP classified abdominal CT findings most reliably when disease was directly expressed as a measurable change in anatomy, attenuation, or burden such as abdominal aortic aneurysm, organomegaly, hepatic steatosis, and atherosclerotic calcification. Performance weakened by design for focal, borderline, or spatial-pattern-dependent findings whose abnormalities fall below what the descriptor space resolves. The phenotype-stratified analysis made this trade-off concrete: CT-IDP fell well below diffuse-subgroup performance in non-calcified, focal, and sub-threshold subgroups, while DINOv3 retained moderate residual discrimination in several of these regimes. CT-IDP and learned image representations occupy partially complementary regimes - anatomy-level descriptors dominating in measurement-defined subgroups, learned representations retaining signal where the abnormality is expressed in spatial pattern, configuration, or local appearance rather than in organ-level summary statistics. Frozen external evaluation reproduced this regime ordering on both the institutional dataset and AMOS, indicating that where CT-IDP is competitive does not depend on dataset identity. AMOS showed larger absolute degradation across all families, consistent with greater protocol heterogeneity particularly contrast phase between AMOS and MERLIN's predominantly portal venous phase distribution.

The interpretability of CT-IDP follows from this design rather than being added on top of it: every prediction is a sparse function of named anatomical measurements, with monotonic, clinically coherent feature-response curves that clinicians can inspect, contest, or audit case by case. The same property makes failures mechanistically inspectable. Failures clustered into recognizable descriptor limitations - focal steatosis missed by a global liver–spleen summary, lesions reported as "too small to characterize" that fell below the segmentation observability floor and the hepatomegaly underperformance on the institutional dataset illustrates the kind of multi-mechanism failure this property is well-positioned to dissect. Coefficient inspection of the trained logistic classifier showed that the third-largest weight was carried by a steatosis-specific descriptor (liver–spleen ΔHU), raising the possibility that NLP-driven training-label structure specifically, preferential co-labeling of hepatomegaly within reports mentioning steatosis produced reliance on a feature whose discriminative value depends on a labeling pattern that is not preserved across datasets. Concurrent contributions from oncologic case mix cannot be ruled out from private institutional data, and a controlled follow-up analysis combining coefficient ablation, stratification by independently validated co-occurring conditions is needed to attribute the external gap quantitatively. The example nevertheless illustrates that descriptor-level interpretability supports the kind of structured diagnostic work-up - coefficient inspection, label-provenance audit, sensitivity analysis that complements, and is harder to organize around, latent embedding models.

CT-IDP differs from much of the traditional radiomics literature, which is predominantly disease-centric: a lesion or organ of interest is segmented, and a frequently texture-dominant feature set is computed within that region for the specific disease under study. CT-IDP instead treats segmented anatomy as a disease-agnostic coordinate system, computing organ- and compartment-level descriptors across the entire abdomen independent of any specific finding. The resulting measurements are best described as image-derived phenotypes rather than validated quantitative imaging biomarkers (QIBs). Advancing selected descriptors toward candidate QIB status will require dedicated validation of bias, precision, repeatability, and reproducibility across vendors, protocols, kernels, dose levels, contrast phases, and segmentation uncertainty; the technical performance agenda established by RSNA's Quantitative Imaging Biomarkers Alliance (QIBA) and continued under Quantitative Imaging Committee. Virtual imaging trials with patient-specific digital phantoms offer one practical route to this validation by allowing controlled perturbation of imaging conditions while preserving known anatomy[28, 29].

The DINOv3 comparison is a contextual benchmark, not an exhaustive deep-learning comparison; the results do not argue that CT-IDP replaces image-based models. Many abdominal findings require representations that capture focal lesions, complex spatial configurations, bowel patterns, or subtle local appearance; exactly the categories where the subgroup analysis placed CT-IDP at or below chance. The natural next step is hybrid: explicit CT-IDP measurements combined with anatomy-aware visual embeddings, each used in its own regime. The same interpretability properties would also support a downstream role in workflow prioritization, in which flagged studies arrive with quantitative evidence supporting the flag, though prospective evaluation would be required to test this.

This study has several limitations. Descriptor fidelity depends on segmentation quality, particularly for small or subtle lesions. Evaluable label sets differed across datasets, and report-derived labels compress radiologically heterogeneous diseases into single binaries, which can obscure operating regimes and produce apparent errors that reflect label derivation rather than model failure. NLP-derived labels additionally exhibit selection patterns such as preferential co-labeling of frequently co-mentioned findings that can introduce spurious feature associations into trained classifiers, as illustrated for hepatomegaly; systematic auditing of such structure is an open methodological question. Per-study contrast-phase metadata was unavailable for AMOS and uncharacterized for the institutional dataset, so the contribution of phase variation to external degradation could not be directly isolated. Sparse logistic regression, while transparent, may under-represent nonlinear interactions. With these caveats, CT-IDP defines its own operating regime: where disease is expressed in measurable anatomy or attenuation, those measurements should be central to the model.

# Conclusion

Across the 8 labels shared by all three datasets, CT-IDP outperformed the baseline on findings defined by organ size, attenuation, or calcific burden, whereas the baseline outperformed CT-IDP on findings defined by compartmental fluid distribution or bowel configuration. Quantitative phenotypes derived from multi-organ segmentation produced cohort-level performance comparable to a vision-transformer baseline across three independent datasets, with consistent gains on findings defined by organ size, attenuation, or burden. Performance degraded predictably for focal and spatial-pattern abnormalities.

*Table 1 Summary of the datasets used in this study. MERLIN served as the development dataset, including training, validation, and internal testing splits. External evaluation was performed on two external datasets: a private institutional clinical dataset and the public AMOS dataset. Labels used indicate the number of evaluable study-level findings included in each dataset for the present analysis.*

| Dataset | Role in study | Patients | Labels used (n) | Age (years) | Female (%) |
|---|---|---|---|---|---|
| **MERLIN** | Development / internal test | 18,321 | 22 | 53.8 ± 19.5 | 55.97 |
| **Private Clinical Dataset** | External Test | 2,000 | 21 | N/A | N/A |
| **AMOS** | External Test | 1,107 | 8 | N/A | N/A |

Table 2 Taxonomy of image-derived phenotypes used for abdominal CT disease classification. Descriptors were grouped into morphometry, attenuation/composition, and contextual/compartmental burden to encode clinically interpretable phenotypes from segmented CT.

| Descriptor family | Representative measurements | Clinical concepts captured |
|---|---|---|
| **Morphometry** | Volume, maximal diameter, surface area, compactness, elongation, and normalized ratios to body habitus | Aneurysm, organomegaly, cardiomegaly, prostatomegaly, pancreatic atrophy, hydronephrosis, and other size or shape-defined abnormalities |
| **Attenuation** | IBSI-aligned first-order HU summaries (for example percentiles and distribution width), cross-organ attenuation contrasts, and constrained high-HU burden within masks | Hepatic steatosis, gallstones, atherosclerotic and valvular calcification, coronary calcification, osteopenia, and other composition-related abnormalities |
| **Contextual / compartmental burden** | Cavity-normalized burden, occupancy, region-specific burden, and anatomy-aware composite ratios | Ascites, pleural effusion, anasarca, renal cyst burden, bowel obstruction, hiatal hernia, free air, and other burden- or compartment-defined abnormalities |

*Table 3* ***Dataset-level benchmark performance of the quantitative phenotyping framework and the DINOv3 baseline.*** *Values are point estimates reported in the main manuscript table. Positive Δ values indicate higher performance of the quantitative model relative to DINOv3.*

| Dataset | Labels | CT-IDP AUC | DINOv3 AUROC | ΔAUC | CT-IDP AP | DINOv3 AP | ΔAP |
|---|---|---|---|---|---|---|---|
| MERLIN | 22 | **0.897** | 0.880 | **+0.017** | **0.713** | 0.685 | **+0.028** |
| External Clinical Dataset | 21 | **0.877** | 0.857 | **+0.020** | **0.760** | 0.735 | **+0.025** |
| AMOS | 8 | **0.780** | 0.756 | **+0.024** | **0.399** | 0.348 | **+0.051** |

*Table 4* ***AUC for the 8 findings shared across MERLIN, and two external clinical datasets – private dataset and AMOS.*** *For each dataset, values shown are AUC for the quantitative phenotyping framework and ΔAUC relative to DINOv3. Positive Δ values indicate higher AUC for the quantitative model.*

| Finding | MERLIN CT-IDP | Δ vs DINOv3 | Private CT-IDP | Δ vs DINOv3 | AMOS CT-IDP | Δ vs DINOv3 |
|---|---|---|---|---|---|---|
| Gallstones | 0.830 | **+0.111** | 0.797 | **+0.255** | 0.771 | **+0.178** |
| Hepatic steatosis | 0.949 | **+0.157** | 0.853 | **+0.064** | 0.836 | **+0.094** |
| Splenomegaly | 0.984 | **+0.016** | 0.980 | **+0.004** | 0.861 | **+0.021** |
| Atherosclerosis | 0.884 | **+0.037** | 0.944 | **+0.014** | 0.781 | **+0.032** |
| Renal cyst | 0.852 | **+0.039** | 0.841 | **+0.027** | 0.561 | **−0.088** |
| Hydronephrosis | 0.783 | **−0.085** | 0.755 | **+0.005** | 0.820 | **+0.039** |
| Ascites | 0.911 | **−0.041** | 0.897 | **−0.070** | 0.848 | **−0.053** |
| Bowel obstruction | 0.878 | **−0.080** | 0.842 | **−0.119** | 0.762 | **−0.028** |

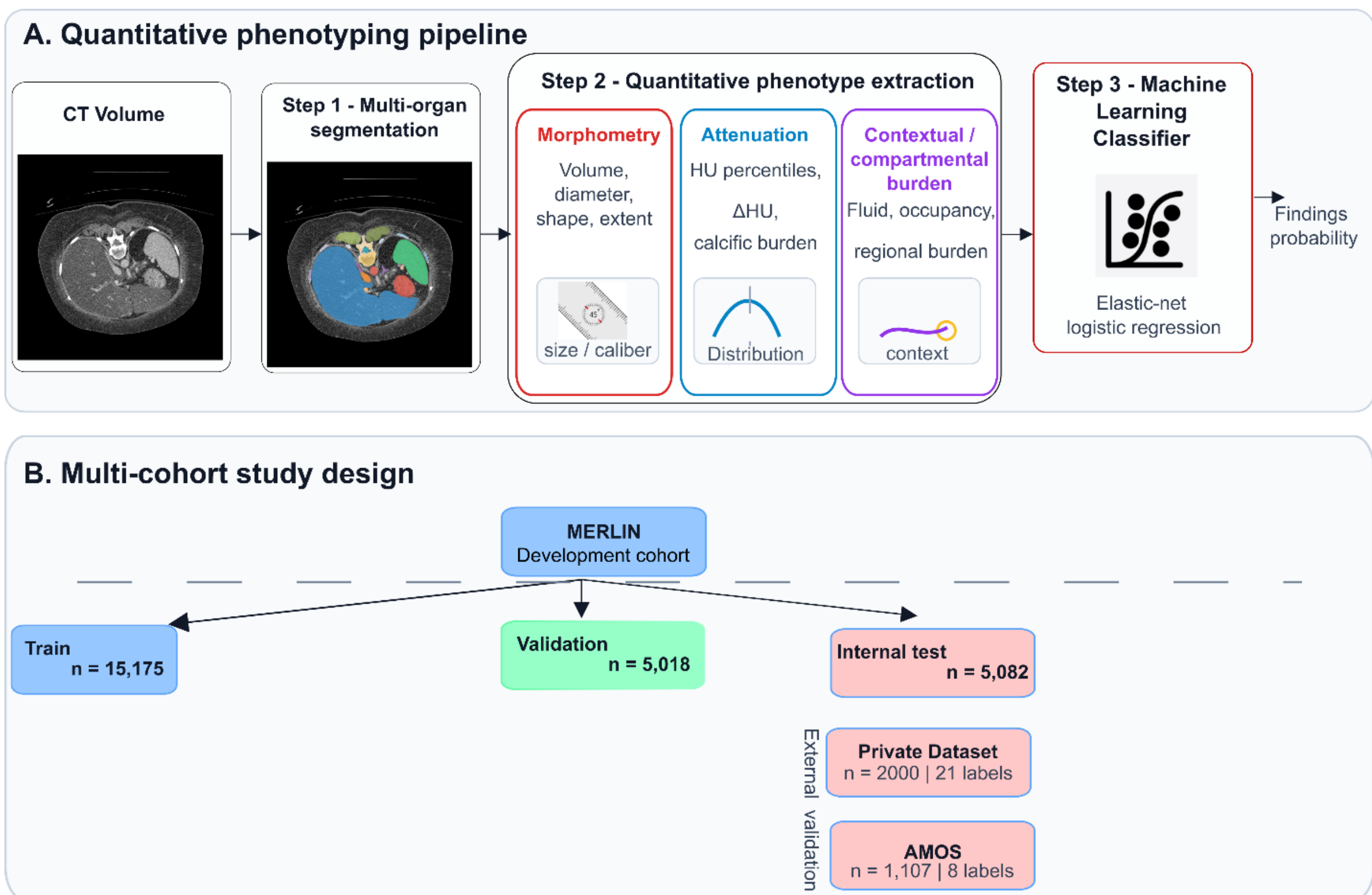

*Figure 1 )* ***Overview of the segmentation-derived quantitative phenotyping framework and study design. (A)*** *Abdominal CT volumes are segmented to establish a whole-body anatomical coordinate system, from which quantitative descriptors are extracted across three families: morphometry, attenuation, and contextual/compartmental burden. These descriptors are used to train sparse disease-specific classifiers that generate interpretable disease probabilities.* ***(B)*** *Study design. MERLIN was used for development and internal testing, and external evaluation was performed on private clinical dataset and AMOS.*

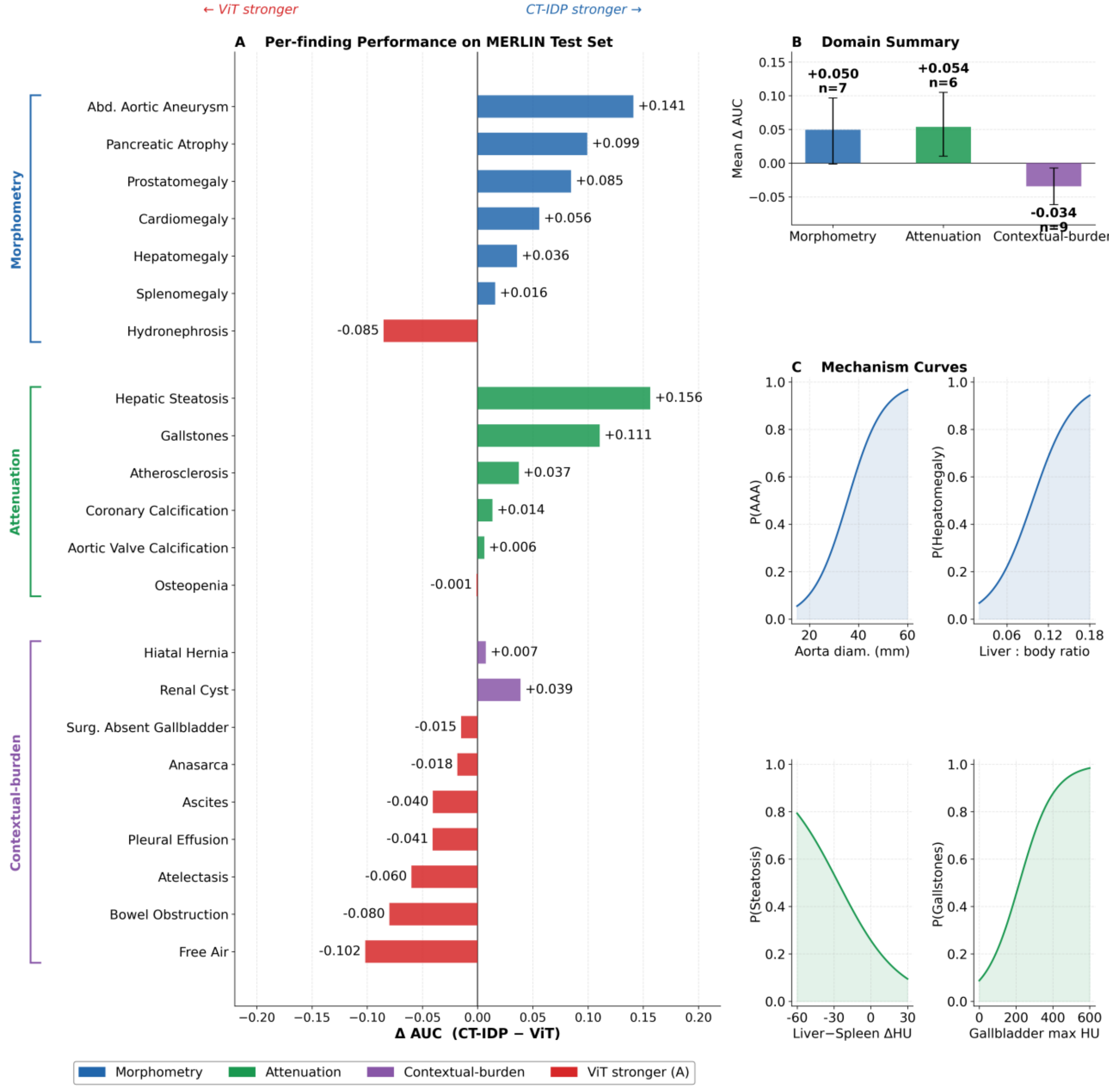

*Figure 2* ***CT-IDP outperforms a ViT baseline on findings defined by measurable anatomy or attenuation and underperforms on spatial-pattern and compartmental findings. (A)*** *Per-finding ΔAUC relative to the DINOv3 baseline across 22 abdominal findings on the MERLIN test set, grouped by descriptor family. Positive values favor CT-IDP.* ***(B)*** *Mean ΔAUC within each descriptor family; morphometry and attenuation are positive overall, contextual burden is negative. Error bars are bootstrap 95% confidence intervals; n denotes the number of findings per family.* ***(C)*** *Ceteris paribus model-response curves for representative retained descriptors, generated by varying one feature across its observed range while holding all others at dataset medians. Curves are monotonic and clinically coherent: predicted probability rises with aortic diameter (AAA), liver-to-body ratio (hepatomegaly), and gallbladder maximum HU (gallstones), and falls with the liver–spleen attenuation difference (steatosis).*

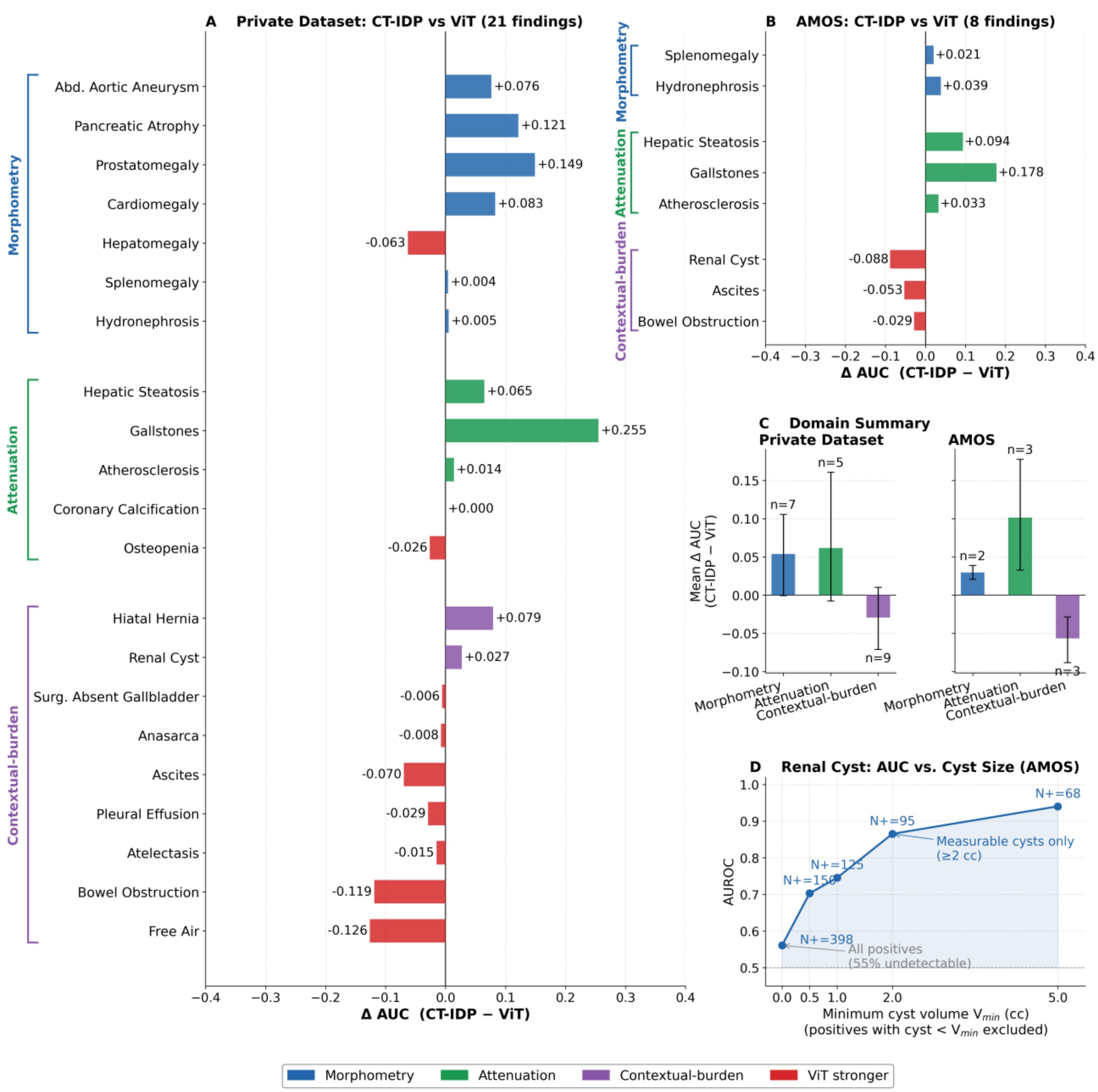

*Figure 3 External validation: per-finding and family-level CT-IDP versus ViT comparison on the private clinical dataset and AMOS. (A) Per-finding ΔAUC (CT-IDP − ViT) on the private dataset across 21 evaluable findings, grouped by descriptor family. Positive (color-coded by family) bars favor CT-IDP; negative (red) bars favor ViT. (B) Corresponding ΔAUC for the 8 findings evaluable on AMOS. (C) Mean ΔAUC by descriptor family on each external dataset; error bars are bootstrap 95% confidence intervals over findings within family; n denotes the number of findings per family. Morphometry and attenuation favor CT-IDP on both datasets; contextual-burden favors ViT. (D) AMOS renal-cyst AUC as a function of minimum cyst-volume threshold $V_{min}$; positive cases with measured cyst volume below $V_{min}$ are excluded. AUC rises monotonically as sub-threshold lesions are removed, indicating that the bottleneck is segmentation observability rather than representation choice. The annotation "55% undetectable" refers to the fraction of report-positive cysts with measured volume below the segmentation observability floor.*

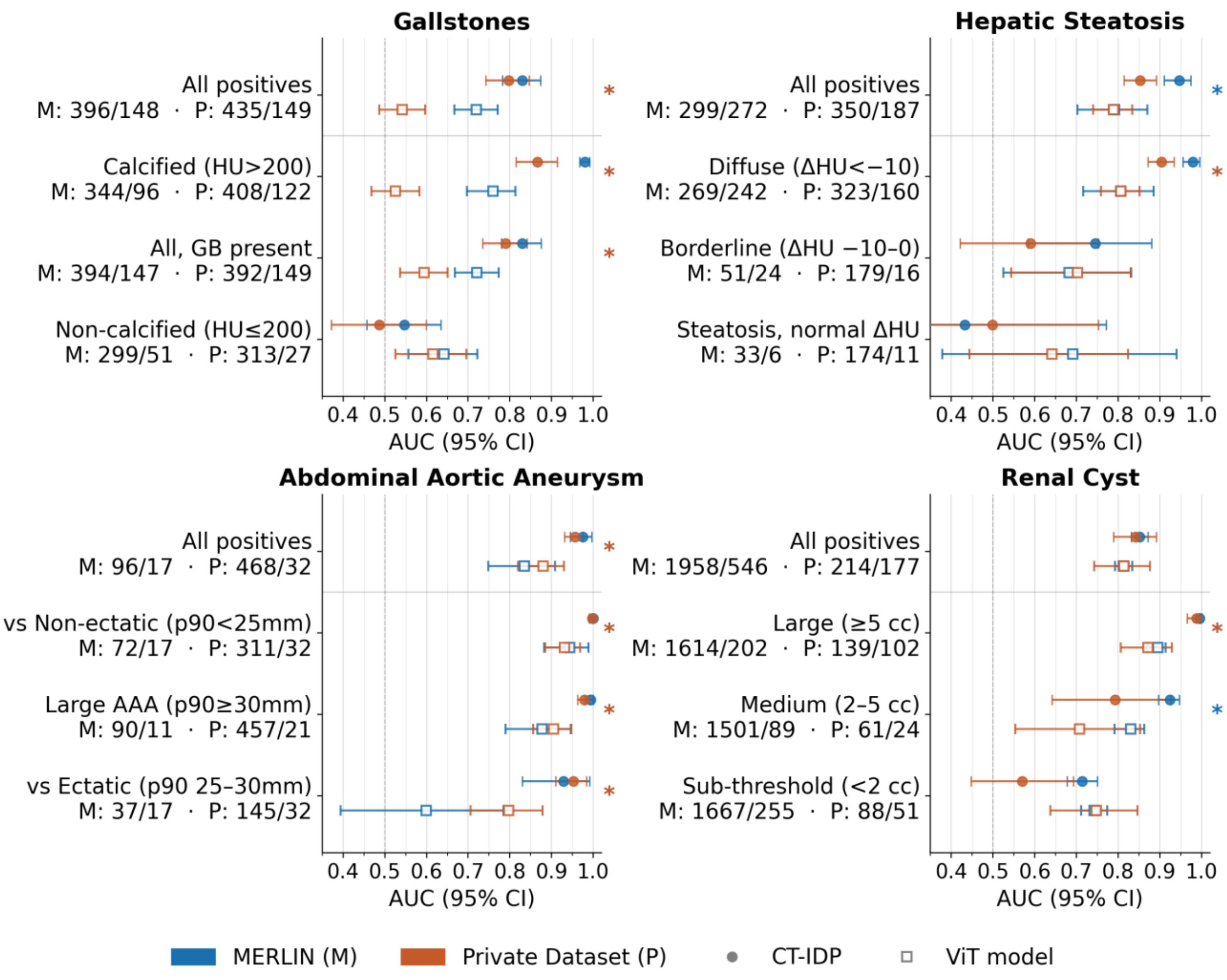

*Figure 4 Phenotype-stratified AUC across four representative abdominal CT findings on MERLIN and the private dataset. For each finding, the full positive dataset is shown alongside subgroups defined by the measurement that operationalizes the diagnosis: gallbladder maximum HU for gallstones, liver–spleen ΔHU for hepatic steatosis, aortic 90th-percentile diameter for abdominal aortic aneurysm, and renal cyst volume for renal cysts. Filled circles denote CT-IDP and open squares denote the DINOv3 baseline; blue and orange encode MERLIN and the private dataset, respectively. Sample sizes are reported as total cases / positive cases per dataset. Asterisks mark strata in which the 95% bootstrap CIs for CT-IDP and DINOv3 do not overlap and are descriptive only; no formal between-model test was applied. Because each stratifying variable is itself a CT-IDP descriptor, results are best interpreted as an observability audit characterizing the regime in which each representation can encode the abnormality, rather than a head-to-head comparison within strata. CT-IDP's discrimination is strongest where disease is well-expressed in segmented anatomy - calcified gallstones, diffuse low-attenuation steatosis, enlarged aortic caliber, and large renal cysts and weakest in non-calcified, focal, sub-threshold, or borderline subgroups, where DINOv3 retains residual discrimination from local appearance cues.*

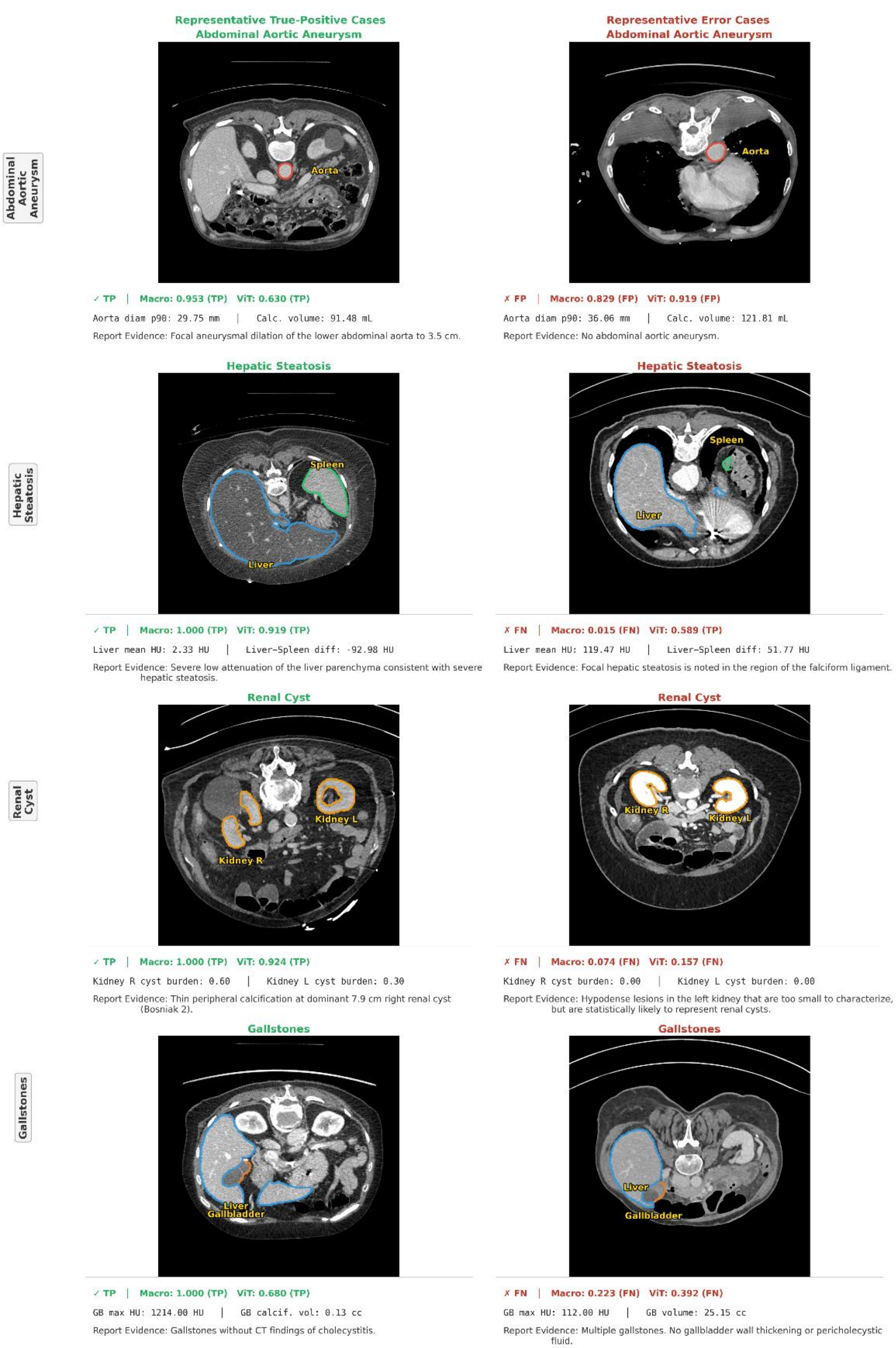

*Figure 5* ***Representative success and failure cases of the CT-IDP framework*** *Cases are shown for abdominal aortic aneurysm, hepatic steatosis, renal cyst, and gallstones. Each panel includes the model outcome, representative quantitative feature values, and a brief report excerpt supporting the reference label. Successes generally occur when disease is clearly expressed in the corresponding quantitative phenotype, whereas failures highlight common limitations, including focal steatosis, sub-threshold renal cysts, and lower-attenuation gallstones.*